# NLP Driven Ensemble Based Automatic Subtitle Generation and Semantic Video Summarization Technique

Aswin VB, Mohammed Javed, Parag Parihar, Aswanth K, Druval CR, Anpam Dagar, Aravinda CV

[1] Indian Institute Of Information Technology Allahabad, Prayagraj, Uttar Pradesh

**Abstract.** This paper proposes an automatic subtitle generation and semantic video summarization technique. The importance of automatic video summarization is vast in the present era of big data. Video summarization helps in efficient storage and also quick surfing of large collection of videos without losing the important ones. The summarization of the videos is done with the help of subtitles which is obtained using several text summarization algorithms. The proposed technique generates the subtitle for videos with/without subtitles using speech recognition and then applies NLP based Text summarization algorithms on the subtitles. The performance of subtitle generation and video summarization is boosted through Ensemble method with two approaches such as Intersection method and Weight based learning method Experimental results reported show the satisfactory performance of the proposed method.

## 1 Introduction

A concept like video summarization has a huge scope in the modern era. Video repository websites like Google, Dailymotion,Vimeo etc. are gaining popularity day by day. The popularity of these websites is enormous in the present scenario. A large amount of online videos are being uploaded as well as downloaded from these online video repository websites. For example, the total number of people who use YouTube are 1,300,000,000[1]. 300 hours of video are uploaded to YouTube every minute! Almost 5 billion videos are watched on YouTube every single day. YouTube gets over 30 million visitors per day[1]. In this scenario, for a concept like video summarization has a huge scope. The video summarization technique can be applied on the video thumbnail to attract more viewers. It can be developed to show only interesting and important parts of the video. It is not necessary for all the videos to come with a subtitle. It is very difficult to summarize the videos like security footage's as they don't have subtitles even after applying speech recognition. This reduces the domain of video summarization. But still summarization of video using subtitles is most efficient and fastest way of doing it. If machine learning algorithms or histogram based methods[2] were used to summarize videos it would have taken a long time to train them which will increase the time of development. But dealing with subtitle which is obvi-ously text is much more easy to deal with and faster which makes the video summarization easier and faster. But the main problem here is that most of the videos comes without subtitles. In such cases, to rectify this problem, the technique of speech recognition which can be applied on the audio of the video and generate subtitles by formatting the text obtained after speech recognition. To extract different sentences from the video there is a need to detect silence in the audio as it recognizes the end of the sentence from that. Once the subtitle is obtained the video can be summarized with the generated sub-titles. For summarization [3] of subtitles Natural Language processing (NLP) algorithms[4] can be used which can be of various accuracy's. Therefore overall this paper proposes NLP based subtitle generation and video summarization technique. Rest of the paper is organized as follows, Section 2 explains the proposed model, Section 3 explains experimental results and Section 4 presents the summary of the report.

## 2      Proposed Model

For summarizing a video using subtitles, the proposed method uses 5 text summarization algorithms which are NLP based methods [5].  Further there is an Ensemble Technique [6] using the Text Summarization [7] algorithms. The Text Summarization algorithms were used for filtering out key contents from Subtitle file (.srt) [8], which will be taken as an input, implement the algorithm on the input, put the sentences in array, and rank them according to their importance using different domains and will pick out the best sentences out of it, so as to form a concise subtitle keeping in mind where those subtitles were originally placed, according to subtitle id.  Next is Ensemble Technique which combines the different algorithms together for a more perfect (intersection of all algorithm) abstract of the input.  Also we are training each algorithm in Ensemble Technique to give precise models. The flowchart in Fig 1 describes the flow of video summarization using subtitles that has been implemented. The input video is fed in along with the subtitles, if the subtitle is not present, subtitle is generated using the subtitle generation algorithm and this subtitle along with the video is used to summarize depending upon whether single algorithm or the combined algorithm is to be used and the summarized video is the output.

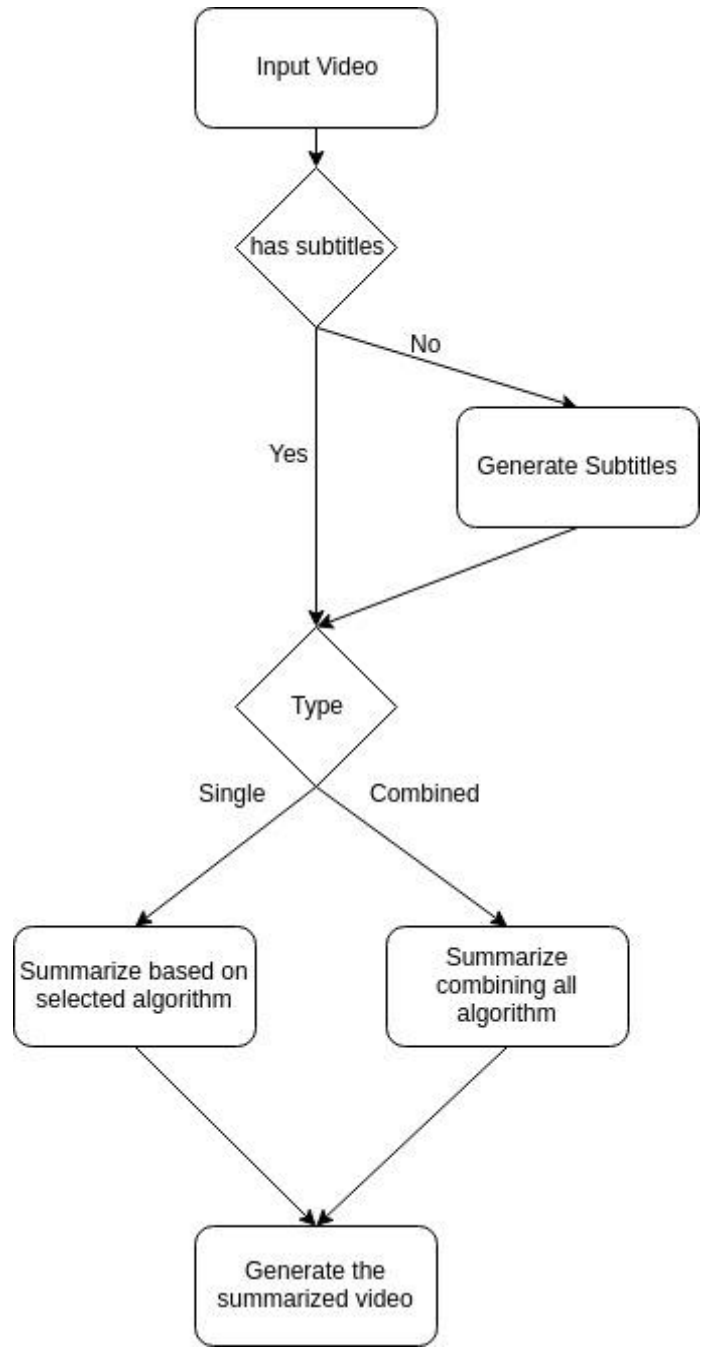

Figure 1. Flow Control of video summarization

## 2.1    NLP Based subtitle generation

Since all videos may not have subtitles along with them and this method can be applied on videos which have subtitles. In case the user does not have a subtitle file then the subtitles is generated first and then process the video using methods listed below. The subtitles (if not provided) are generated using Speech recognition API of WIT.AI [9] which is used by Facebook for speech recognition. Basic idea on how the subtitle is generated is that chunks of audio is extracted from the video and apply speech recognition on them. To elaborate this, First the audio is extracted from the video file. Then max time interval is fixed for each subtitle for the video let it be 6 sec which is in our case. Then the audio is scanned to detect silence in it if the silence occurs before 6 seconds the audio will be cut at that point. Also a threshold is defined such that above this level the part of the audio is not treated as silence and vice versa. Also, to be noted that 1 second of extra silence is added at the starting and at the ending of the audio. So that words to be recognized by the speech recognizer is not missed out. Each time the words are recognized in a particular chunk of audio the whole sentence is formatted in the form of a subtitle file such that each of the sentence will be mentioned with the starting time stamp and the ending time stamp. Once the subtitle is available it can be summarized to obtain the summarized video. There are 5 Text Summarization algorithms which have been used to summarize a video. These are:

### 2.1.1    Luhn

1958 came around as an emerging year in the field of summarization, when Hans Peter Luhn [9] suggested an algorithm for thought summarization. This was a carrying a lot of weight achievement and a big head start in this sector, and followed his consider was a action in summarization area. Luhn received a rule of thumb to recognize salient sentences from the text by features well known as definition and definition frequency. This algorithm checks for words that have high occurrence frequency and the words are sorted based on decreasing frequency. The weight of a sentence is calculated by summing weights of each relevant word and these sentences are sorted in decreasing order based on the summed weight and finally P most relevant sentences are taken from the subtitles as the output [10].

### 2.1.2    LSA

Latent Semantic analysis (Also known as LSA) [10] is based on words and concepts. If each word represent only a single concept LSA would have been lot easier but un-fortunately each word in English language represent several concepts (like synonyms).Because of this varying concepts within each word LSA becomes a little tricky as each word map to more than one concept. The basic concept of LSA is based on Title words and Index words. Title words are those words which appear in all the sentences in LSA and Index words are those words which appear in more that 2 sentences which has the Title word. A document is considered as a bag of words. The order in which the words are arranged is not important whereas the count of a particular word plays an important role. Also, each word is suppose to have only one meaning. LSA is based on concepts which are represented as a pattern of words. To understand the concept behind the word clustering is used in LSA. The basic idea is to plot a XY graph including all the index words and title words based on their occurrences in each sentences. Then all the clusters in the graph are identified. Each of these clusters represent each concept and title in that sentence represent that particular concept. Hence concept for each title word can be extracted. So in summarization, the technique was used in such a way that the sentence which is included in the most crowded cluster of the graph was taken. Using this concept of Title words and Index words any text document can be summarized [11].

### 2.1.3 Text Rank

In the Text Rank algorithm [12], it first takes the text which has to be split up and converted into sentences and further into vectors. A similarity matrix is constructed from the vectors and a graph is created from this matrix. Using this graph, the sentence ranking is done. Based on the ranked sentences, summarized text is obtained. The probability of going from sentence A to sentence B is the similarity of 2 sentences. The text units which best defines the sentence are identified. These text units are then added to the graph as vertices. The relations which connect texts are identified which are used to draw edges of the graph between vertices. A graph based ranking algorithm is used until it converges. During the ranking algorithm, each vertex is assigned a value, this value is used for se-lection decisions. Finally the vertices are sorted on the basis of their final score value and then the sentences are sorted based on the sorted vertices in the graph.

### 2.1.4 Lex Rank

In the Lex Rank algorithm [13], first all the nouns and adjectives is separated out to form a document cluster. Now the IDF (Inverse Document Frequency) scores is found for each words in this cluster. For each word let tf be the frequency of that word in the cluster. The collection of all words which have $t_f * IDF$ score greater than a threshold value forms the centroid of the cluster. So importance of a sentence will be higher if it contains more number of words which are present in the centroid. So using this concept, P most relevant sentences are selected [13].

### 2.1.5 Edmundson

The previous works of Luhn showed that the key words can be extracted out as the most frequently occurring content words except the stop words and summarize the text according containing the most key words. But since this much is not alone capable to generate proper summarized text,

Edmundson [14] proposed another algorithm and added three more methods to extract out key words, which are namely, pragmatic words (cue words),title and heading words, and structural indicators (sentence location). The results of Edmundson cleared out the fact that the other three factors played a dominant role in generating the best summarized text (as expected). Since it uses extra words call stigma and bonus word, the summarization will be biased according to those files. So this method will not be used in the proposed method [14].

## 2.2   Video Summarization

From the algorithms explained in the previous section, Edmundson Summarization will not be used because it is biased according to the bonus and stigma words given by the user. So it cannot be used for a comparison.

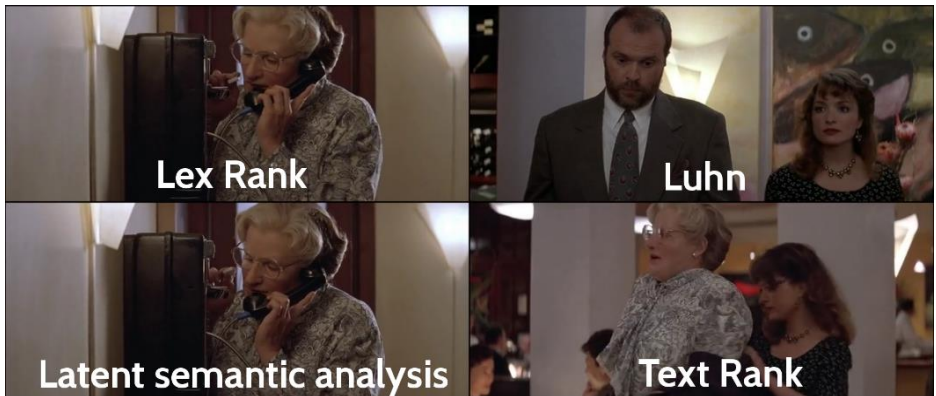

Figure 2. Output of Summarization using 4 algorithms.

Fig 2 is the summarization of the movie named Mrs.Doubtfire, contains a frame of output using 4 Algorithms (Lex Rank, Luhn, LSA and Text Rank). A video of Shah Rukh Khan's Ted Talk which was of 17 minutes was taken and summarized. The subtitle file of the input video had 370 lines. The video was fed into the 4 above mentioned algorithms separately, the graph of subtitle text with it being relevant in the summarized video for the algorithms are described in Fig 3, Fig 4, Fig 5, Fig 6 .

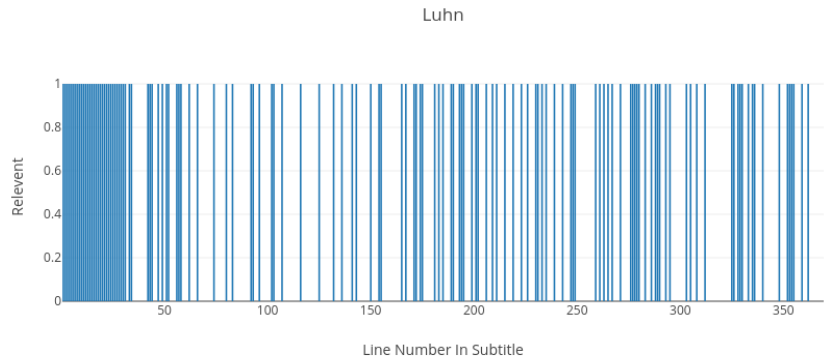

Figure 3. Luhn Algorithm.

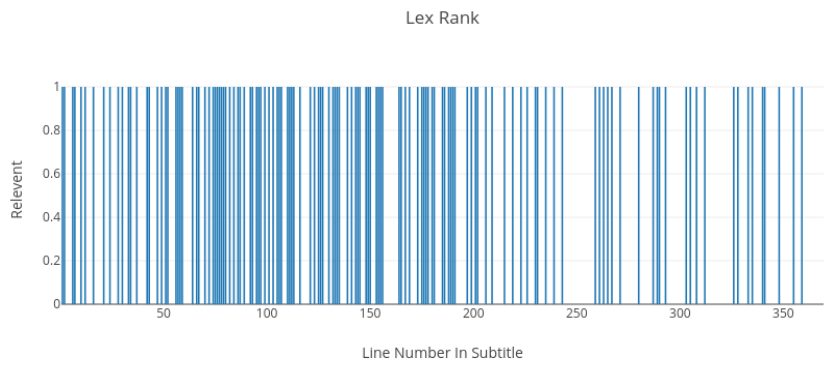

Figure 4. Lex Rank Algorithm

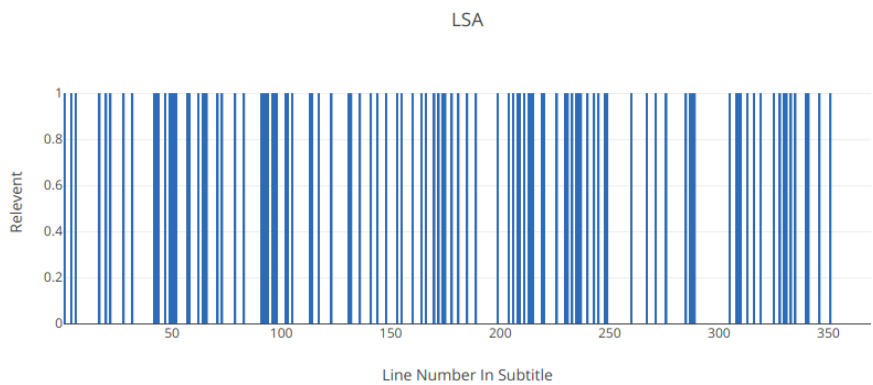

Figure 5. LSA Algorithm.

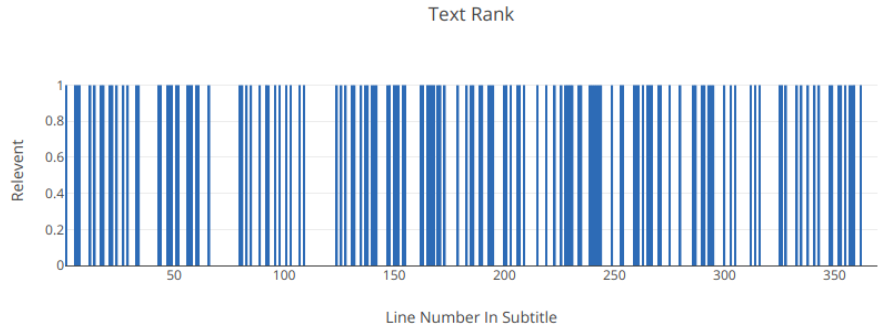

Figure 6. Text Rank Algorithm

### 2.2.1 Taking Intersection and Combining

A very basic and simple approach is to combine them directly, meaning, the idea is to run all the algorithms, keep their outputs side by side, and take a simple mathematical intersection. As it is very obvious, if a sentence is in all the combining algorithms, then its must be of great importance, hence it should be included in the output file. Hence, the user can choose from Luhn, LSA, LexRank, TextRank, which all he wants to combine multiple correct, so that he can get the best results when all of them combined. Also, Edmundson cannot be concatenated in this process because it requires two extra files of Bonus Words and Stigma Words, hence, taking that while combining them was not ideally possible. This Intersection gave very good results and implemented the ideology of Ensemble properly.

### 2.2.2 Weight Based Learning Algorithms

The problem with the previous method was that, it gave equal powers to all the algorithms, but, from the above explanations it can be seen that not all algorithms behave properly, so, here a method was devised, where each could get some weightage. The idea was simple, 'The one that performs the best, gets more'. As the name suggest, an initial weights were taken for all algorithm, and initialized all the algorithms with the same weights, ie,WL, WLSA, WE, WLR, WTR = 1.Now, the check function will compare output of each of the algorithms and rank them accordingly on basis of their performance, so the ones that performed may get an increment in their weights. So, during future summarizations, they will get their scores according to their weights, and the sentences with higher score, will be part of the output file. In this way a clear view can be obtained of which algorithm can give best output for the given input and the suggested algorithm can be used to obtain better results. The figure 7 is the weight allocation for different algorithms before and after summarization of Shah Rukh Khan's video. From the figure it can be understood that the weight of LSA was increased and that of Lex was decreased. So for this video LSA performed better and Lex performedthe least.So the weight of Lex Rank is increased by a unit and that of Luhn is decreased.

Figure 7. Weights before and after summarization of first video.

## 3    Experimental Results

Since there is no dataset for video summarization using subtitles and the 4 summarization technique, a dataset of 40 videos was generated which were having different time length. These videos were given to these 4 algorithms and also the combined algorithm and the number of lines obtained in the summarized video was noted. The efficiency of each algorithm was decided based on the output of combined video which was the intersection of all these algorithms. So the efficiency can be defined for an algorithm as the ratio of the number of subtitles in the combined video to that of the output of the particular algorithm.

$$efficiency = N_{combined} \: / \: N_{algorithm} \qquad (1)$$

### 3.1    Efficiency of Video Summarization
By efficiency of an algorithm in summarization, it is meant how much part of the summarized algorithm is present in the video generated by the ensemble technique. As mentioned above, efficiency of each algorithm can be calculated using formula 1 and on applying this on the intersection method following results were obtained as shown in table1.

**Table 1.** Efficiency of Intersection Method.

| Lex Rank | LSA | Luhn | Text |
|----------|------|------|------|
| 37.1 | 40.6 | 38.0 | 37.7 |

On applying the dataset on the weighted ensemble technique, with initial weights set to 1 for all 4 algorithms, the following results were obtained as shown in table 2 and the updated weights are shown in table 3.

**Table 2.** Efficiency of Weighted Ensemble Method.

| Lex Rank | LSA | Luhn | Text |
|----------|------|------|------|
| 18.5 | 61.0 | 38.0 | 37.7 |

**Table 3.** Weights Ensemble Method.

|  | LSA | Luhn | Text | Lex Rank |
|--------|-------|------|------|----------|
| **Initial** | 1 | 1 | 1 | 1 |
| **Finial** | 1.975 | 1 | 1 | 0.025 |

From the tables 1 and 2 it is observed that LSA performed better and Lex Rank performed the least. Since for all videos LSA performed best and Lex performed least their weights got affected whereas the weight of Luhn and Text Rank remained unchanged (Table 3). There is a huge difference in the efficiency value in weight based and intersection method because in weight based ensemble technique at each iteration the weight of the better algorithm is increased and that of the worst algorithm is decreased. From these two methods it is clear that LSA has a major contribution to the summarized video and Lex has the least contribution.

### 3.2 Complexity

**Time Complexity**

1. Single Summarization Algorithm: $O(nk)$
   where n is the number of iterations until the summarization length is obtained and k is the number of sentences in the summarized subtitles.
2. Combined Summarization Algorithm: $O(\sum_{i=1}^{\alpha} n_i k_i + \min(k_i))$
   where $\alpha$ is the number of methods to be combined, n is the number of iterations until the summarization length is obtained, k is the number of sentences in the summarized subtitles.

**Space Complexity**

1. Single Summarization Algorithm: $O(r + L * r)$
   where r is the total number of regions in the subtitle array, L is the average length of the sentences in the summarized subtitle.

2. Combined Summarization Algorithm: $\mathbf{O(\sum_{k=1}^{\alpha} r + L * r)}$
   where r is the total number of regions in the subtitle array, L is the average length of the sentences in the summarized subtitle.

## 4      Conclusion

Large number of videos are being generated and are increasing day by day. Hence, video summarization technique will be very helpful. Video summarization provided a faster way browsing of large video collections and more efficient content indexing and access. The use of NLP Algorithms proved to be a very efficient way to form abstracts of videos. The case of no subtitles was by using subtitle generation method to convert speech to text, which turned out to be of great use in normal day to day usage. Many of the videos which is taken from phones, etc., do not contain subtitles, hence there is future scope to work on this problem.